\documentclass[conference]{IEEEtran}
\IEEEoverridecommandlockouts

\usepackage{amsmath,amssymb,amsfonts}
\usepackage{algorithmic}
\usepackage{graphicx}
\usepackage{textcomp}
\usepackage{xcolor}

\usepackage{booktabs} 
\usepackage{microtype}
\usepackage{todonotes}
\usepackage{url}
\usepackage{amsmath}
\usepackage{amssymb}
\usepackage{placeins}

\usepackage{subcaption}

\usepackage{multirow}
\usepackage[numbers]{natbib}
\begin{document}

\title{Learning Local Forward Models\\ on Unforgiving Games}


\author{Alexander Dockhorn${}^*$, Simon M. Lucas${}^+$, Vanessa Volz${}^+$, Ivan Bravi${}^+$, Raluca D. Gaina${}^+$, Diego Perez-Liebana${}^+$
    \thanks{${}^*$ with the Computational Intelligence Research Group, Otto von Guericke University, Magdeburg, Germany.}
    \thanks{${}^+$ with the School of Electrical Engineering and Computer Engineering, Queen Mary University of London, London, UK.}
}

\IEEEpubid{\begin{minipage}{\textwidth}\ \\[12pt]
978-1-7281-1884-0/19/\$31.00 \copyright 2019 IEEE
\end{minipage}}

\maketitle


%
\begin{abstract}
This paper examines learning approaches for forward models based on local cell transition functions. We provide a formal definition of local forward models for which we propose two basic learning approaches. Our analysis is based on the game Sokoban, where a wrong action can lead to an unsolvable game state. Therefore, an accurate prediction of an  action's resulting state is necessary to avoid this scenario.

In contrast to learning the complete state transition function, local forward models allow extracting multiple training examples from a single state transition. In this way, the Hash Set model, as well as the Decision Tree model, quickly learn to predict upcoming state transitions of both the training and the test set. Applying the model using a statistical forward planner showed that the best models can be used to satisfying degree even in cases in which the test levels have not yet been seen.

Our evaluation includes an analysis of various local neighbourhood patterns and sizes to test the learners' capabilities in case too few or too many attributes are extracted, of which the latter has shown do degrade the performance of the model learner.
\end{abstract}

\begin{IEEEkeywords} Forward Model Learning, Local Forward Model, Decision Tree, Rolling Horizon Evolutionary Algorithm
\end{IEEEkeywords}

%
%
%


\section{Introduction}

Learning Forward Models (FMs) is an important challenge in Artificial Intelligence (AI). An FM is used to simulate future system states given an initial state and a sequence of actions. FMs are essential for Statistical Forward Planning (SFP) methods, such as Monte Carlo Tree Search or Rolling Horizon Evolutionary Algorithms. Learning FMs is an active subject of study involving several approaches, notably deep learning and rule induction.

In previous work \cite{Dockhorn2018,Lucas2019LocalFML}, we proposed to use an ensemble of local models as an FM in grid-based games. In these, the next state of a given cell often depends on the surrounding ones and thus seem specifically well suited for this approach. 
Thus, by assuming locality, the amount of data required to train reliable FMs can be reduced in comparison to end-to-end models. Usually, this assumption is justifiable, as locality can often be easily identified by studying the game's ruleset.

The proposed approach was shown to work well on several games of the GVGAI framework \cite{Dockhorn2018} and the \emph{Game of Life} \cite{Lucas2019LocalFML}. However, the performance of an AI in the Game of Life has shown to be robust to a small number of errors in the FM, as decisions based on wrong predictions usually hamper the performance only temporarily. In addition, even one game tick provides a large number of training samples.

In this paper, we thus test the local FM approach further on the puzzle game Sokoban which usually has a smaller map size than the \emph{Game of Life} and contains a larger number of tile types. FMs thus need to be trained on smaller data sets. In addition, Sokoban is an \emph{unforgiving} game. This means that a single wrong decision can at times lead to an irreversible state that results in losing the game. This aspect translates to an added layer of difficulty for training the FM, as the prediction accuracy for specific states is crucial to winning the game.

In the following, we first provide a description of Sokoban and the difficulties associated with the game. In section \mbox{\ref{sec:methods}}, we give an overview of the local modelling approach used in this paper. We describe our experimental setup in \mbox{section \ref{sec:exp_setup}} and our results in section \ref{sec:results}. We draw conclusions from our findings in section \ref{sec:conclusion} and discuss promising future work.




\section{Sokoban}
\label{sec:sokoban}

\begin{figure}[!t]
	\begin{center}
	\includegraphics[width = .15\textwidth]{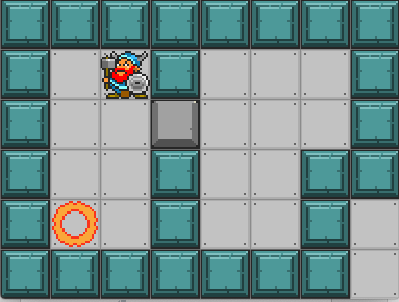}
	\includegraphics[width = .15\textwidth]{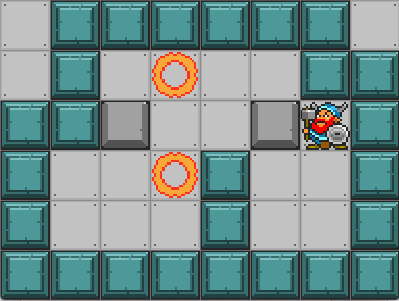}
	\includegraphics[width = .15\textwidth]{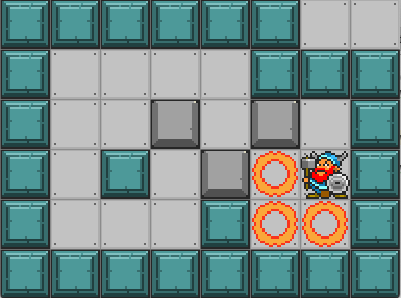}
	\caption{Examples of Sokoban levels. The player (dwarf avatar) must push all boxes (grey blocks) into the targets (orange circles) to complete the level.}
	\label{fig:sokoban}
	\end{center}
\end{figure}

Sokoban (Thinking Rabbit, 1982) is a classic puzzle game in which the player must push a determined number of crates into designated locations to complete each level. The player can move in four directions (up, down, left, and right) and push boxes in the direction of travel. Pulling boxes is not possible. The levels in Sokoban contain immovable blocks and the play area is surrounded by walls. Figure~\ref{fig:sokoban} shows three examples of levels for this game. There are thousands of Sokoban levels available\footnote{http://www.sourcecode.se/sokoban/levels} created by the community around this game.

The main complexity of Sokoban resides in the fact that it has trap moves: pushing a crate against a wall may importantly limit the positions to which this box can be moved further, possibly making the level unsolvable. Additionally, the game is difficult to solve for game playing AI agents, as its rewards are sparse. By default, the only feedback is the victory condition when all crates are placed at the predefined locations or number of boxes on goal tiles. In order to smooth out the reward landscape, we will make use of the latter. The combination of these two aspects makes Sokoban an interesting benchmark for AI and, in particular, for FM learning. 

The implementation of Sokoban used for this paper can be found in a GitHub repository\footnote{\url{https://github.com/GAIGResearch/LearningFM}}. This is an efficient implementation of the game (running approximately at 12.5 thousand ticks per second\footnote{using a MacBook Air, 1.8 GHz Intel Core i5, 8 GB 1600 MHz DDR3}) written in Java and Kotlin.


\section{Methods}
\label{sec:methods}

\subsection{Learning Local Forward Models}

While learning an FM is a hard problem in general,
it can be made more approachable by considering only local interactions
(or perhaps \emph{mostly} local interactions). 
This approach has been motivated by the experiments of \cite{dockhorn_2018} in which local models of the player and its surrounding objects were used to predict future states of the game and their rewards.

\subsubsection{Definition of a Local Model}
Given a system (game) state $S_t$ perceived through multiple sensors $(S_t^{(1)}, S_t^{(2)}, \dots, S_t^{(n)})$, a state transition function $f$
maps the current state and the agent's action $A_t \in \mathcal{A}$ to the next system state $S_{t+1}$
\begin{equation}
    f:  \mathcal{S}, \mathcal{A} \rightarrow \mathcal{S}
    \qquad S_t, A_t \longmapsto S_{t+1}
\end{equation}

In contrast to mapping the transition of the whole state and therefore the transition of all its sensor values in a single transition function, the game's transition function can also be split into multiple components, each mapping the transition of a subset of observable values or a single value:
\begin{equation}
    f_i:  \mathcal{S}, \mathcal{A} \rightarrow \mathcal{S}^{(i)}
    \qquad S_t, A_t \longmapsto S_{t+1}^{(i)}
\end{equation}

This approach can be useful when the transition function for each observable value can be reduced due to its independence of some of the state's components.
Especially in grid-based games, this characteristic is represented by the absence of global effects such that the future state of a cell can be determined by only observing the local neighbourhood of the cell.

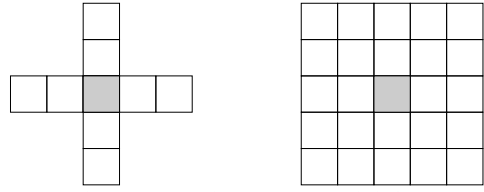
\begin{figure}
    \centering
    \resizebox{0.35\textwidth}{!}{%
    \begin{tikzpicture}
        
        \foreach \x in {-2,...,2} {
            \node[draw, rectangle, minimum width=1.0cm,minimum height = 1.0cm] at (\x,0) {};
        }
        \foreach \y in {-2,...,2} {
            \node[draw, rectangle, minimum width=1.0cm,minimum height = 1.0cm] at (0,\y) {};
        }
        \node[draw, fill=black!20, rectangle, minimum width=1.0cm,minimum height = 1.0cm] at (0,0) {};
        
        \foreach \x in {-2,...,2} {
            \foreach \y in {-2,...,2} {
                \node[draw, rectangle, minimum width=1.0cm,minimum height = 1.0cm] at (8+\x,\y) {};
            }
        }
        \node[draw, fill=black!20, rectangle, minimum width=1.0cm,minimum height = 1.0cm] at (8,0) {};
       
    \end{tikzpicture}
    }

    \caption{Local neighbourhood patterns (which include the centre tile) used to predict the next state of the centre tile.}
    \label{fig:pattern_types}
\end{figure}

In this paper, we restrict ourselves to the consideration of models for grid-based games. In this games, a state can be represented as a set of tiles arranged
in a grid where $T(x,y)$ specifies the tile at position $x,y$ on the grid.
For all grid tiles we aim to predict the state of each tile at time
$t+1$ based on the state of the cell and its local neighbourhood at time
$t$. Let the local state transition function $f_{x,y}$ be given by:
\begin{equation}
    f_{x,y}:  \textit{N}(x,y)_t, A_t \longmapsto  T(x,y)_{t+1}
\end{equation}
In this work we consider two types of local neighbourhoods, namely the cross pattern and the square pattern (cf. \ref{fig:pattern_types}).
\begin{equation}
    \begin{split}
        \textit{N}_{\textit{cross}}(x,y) = \big\lbrace T(x+i,y) ~&\big\vert~  0\leq \vert i\vert \leq span \big\rbrace~ \cup\\
         \big\lbrace T(x,y+j) ~&\big\vert~ 0\leq \vert j \vert \leq span\big\rbrace
    \end{split}
\end{equation}
\begin{equation}
    \begin{split}
        \textit{N}_{\textit{square}}(x,y) = \big\lbrace T(x+i,y+j) ~\big\vert~  &0\leq \vert i\vert \leq span,~ \\
        & 0\leq \vert j \vert \leq span\big\rbrace
    \end{split}
\end{equation}
Various spans will be tested to explore the influence of providing the learning approaches with too much or not enough information to predict the centre tile.

Access to the previous history may also be possible, but
whether or not this is necessary depends
on whether the game (or our observation of it)
is Markovian.\footnote{A system is Markovian if its transitions can be described by a memoryless stochastic process.}  
Whether Sokoban is Markovian depends on how we observe it.
If we are just observing the tile image, it is non-Markovian,
since an avatar may be on top of an empty space or a hole, 
and just by observing the current tile image there is no
way to tell the two possibilities apart. This has been overcome previously by 
having a separate input plane for each tile type \cite{DBLP:journals/corr/WeberRRBGRBVHLP17}.  Here, we
solve it in a more space-efficient way by using a different symbol for the avatar depending on whether it is over a space or over a hole.

\subsection{Implementation of Local Forward Models}

In this work we consider two algorithms for local forward model learning, the \textit{Hash Map} and the \textit{Decision Tree Model}.

\paragraph{Hash Map Model}
After a state transition is observed, the Hash Map Model extracts the action and the cell values specified by the neighbourhood pattern.
For each cell in the grid, it stores the extracted values and the observed future state distribution of the centre cell as key-value pairs in a Hash Map. Note that we store a distribution rather than a single tile, since when local models capture insufficient context the same local pattern may map to many possible next centre cell values.

The generated Hash Map can be used to predict transitions to upcoming states given the current grid and an (anticipated) action.
A future state can be generated by once again extracting the local neighbourhood of each cell and looking up the most likely future state of the centre cell in the Hash Map.
In case the Hash Map does not contain a given key, we assume that no change of the centre cell will occur. This is a good baseline assumption, as most cells do not change in an iteration.

\paragraph{Decision Tree Model}
Similar to the Hash Map Model, we first generate a training set of observed transitions per cell by extracting tuples of action, local neighbourhood cell states, and the resulting state of the centre cell.
Using this training set, a Decision Tree is trained to create a mapping from the specified input to the future cell state.  
In contrast to the Hash Map Model, the Decision Tree is able to generalise from provided examples and can be used to classify unobserved input patterns.  However, unlike our Hash Map Model, the Decision Tree currently stores only the first cell type to occur for each local pattern, not a distribution of cell types.
We chose to create an unpruned Decision Tree to not exclude FM rules that only occur rarely.

\subsection{Agent Model}

We test the applicability of learned FM using a Rolling Horizon Evolution (RHEA) agent based on a $1+1$~EA. This agent generates a new solution (action sequence) at every iteration through mutation of the original and replaces the original if the new solution has a higher or equal fitness. Each solution is evaluated by using the game's model to simulate all actions in the sequence, and the resulting game state is assessed by a heuristic function (the game's score, which, in the case of Sokoban, is the number of boxes placed on targets). The sequence may be evaluated several times in order to reduce the noise, and the average of all final-state values is used as the solution's fitness. RHEA has proven effective across a range of games \cite{lucas2019efficient, lucas2018n, gaina2017rolling}.
A detailed description of the agent can be found in \cite{Lucas2019LocalFML}.
The FM of the game is key to the evolution process (and thus the performance of the agent), as it is used to evaluate the fitness of generated solutions. Inaccurate models could, therefore, lead to erroneous estimations of solution outcomes (and implicitly their values).

\section{Evaluation of Model Accuracy and Playing Performance}
\label{sec:exp_setup}

For the experimental evaluation of the proposed local FM learning methods, we trained the Hash Map and the Decision Tree model on 10 Sokoban levels and tested their performance in two regards - (1) achieved prediction accuracy and (2) its effects on the performance of a game AI.

For the first set of experiments, each model is trained using observed state transitions of a randomly moving agent.
Each of the ten training levels was played 100 times for 100 game steps each. Then, we collected test data from unseen levels in the same fashion. We use two test level sets of which the first contains a single easy level and the second consists of 10 more complex levels. Both data sets thus contain independent sets of observed transitions. 

For each state transition in the complex test set, we let the trained models predict the next state using the previous game state and the agent's action.
The predicted grid is compared to the real game state observed.
We measure the accuracy of the one-step prediction by counting the number of correctly predicted tiles and dividing it by the number of total tiles.

In a second set of experiments (resulting playing performance), the same pre-trained models are used in conjunction with the RHEA agent to play the 10 unseen test levels.
Using the N-Tuple Bandit Evolutionary Algorithm (NTBEA) \cite{lucas2018n} we first optimised the agent's sequence parameters (selected values shown in parentheses) including sequence length (40), number of evaluations (40), mutation rate (0.4) and shift buffer (true) by evaluating its playing performance on the training levels.
The final agent was used for evaluating the playing performance while using either the true model, learned models, or a static baseline model (predicting no change in the game state).

In the evaluation of the models' accuracy and the models' resulting game playing performance, we varied the pattern of the local neighbourhood and its span to determine their influence on both performance measures.
Theoretically, the perfect model would require a span of 2 (i.e. a 5x5 square grid or a cross within a 5x5 grid) to correctly model interactions that involve the agent pushing a block.
Agent movement not considering boxes can be predicted using a span of 1.
Wall tiles can be predicted without knowledge of their local neighbourhood since they will remain static.

 \begin{table}[!t]
        \centering
        \begin{tabular}{cccc}
            \toprule
                \multirow{2}{*}{model}& easy & hard \\
                 & score & score \\
            \midrule
                Static Forward Model & 0.40 & 0.64  \\
                True Forward Model &  2.96 & 0.89 \\
            \bottomrule
        \end{tabular}
        \caption{Baseline results}
        \label{tab:baseline}
    \end{table}
    
    \begin{table}[!t]
        \centering
        \begin{tabular}{cc|c|cc|cc}
            \toprule
            &\multirow{2}{*}{span} & unique & \multicolumn{2}{c|}{easy} & \multicolumn{2}{c}{hard} \\
            & &  ~~patterns~~ & acc & score & acc & score\\
            \midrule
            \multirow{3}{*}{\rotatebox{90}{cross\vphantom{q}}}&1 & 5000& 0.9930 & 0.73 & $0.9965$ & $0.64$\\
            &2 & 27419& 0.9799 & 0.42 & $0.9894$ & $0.65$\\
            &3 & 46271& 0.9771 & 0.41 & $0.9869$ & $0.65$\\
            \midrule
            \multirow{3}{*}{\rotatebox{90}{square}} &1 & 27667& 0.9822 & 0.35 &  $0.9919$ & $0.65$\\
            &2 & 151995& 0.9773 & 0.43 &  $0.9864$ & $0.64$\\
            &3 & 303200& 0.9770 & 0.46 &  $0.9863$ & $0.65$\\
            \bottomrule
        \end{tabular}
        \caption{Results of the \textit{Hash Map Model}}
        \label{tab:hashmap}
    \end{table}
    
    \begin{table}[!t]
        \centering
        \begin{tabular}{cc|c|cc|cc}
            \toprule
            & \multirow{2}{*}{span} & nr of & \multicolumn{2}{c|}{easy} & \multicolumn{2}{c}{hard} \\
            & &  \,tree nodes\, & acc & score & acc & score\\
            \midrule
            \multirow{3}{*}{\rotatebox{90}{cross\vphantom{q}}}&1 & 679& 0.9959 & 0.00 & $0.9973$ & $0.67$\\
            &2 & 459& 0.9991 & 1.11&  $0.9997$ & $0.80$\\
            &3 & 664& 0.9990 & 0.75&  $0.9995$ & $0.78$\\
            \midrule
            \multirow{3}{*}{\rotatebox{90}{square}}&1 & 1638& 0.9972 & 0.00 &  $0.9975$ & $0.63$\\
            &2 & 2088& 0.9975 & 1.06 & $0.9988$ & $0.63$\\
            &3 & 2864& 0.9981 & 0.66 &  $0.9985$ & $0.67$\\
            \bottomrule
        \end{tabular}
        \caption{Results of the \textit{Decision Tree Model}}
        \label{tab:decisiontree}
    \end{table}

\section{Results}
\label{sec:results}

The results of our experiments are summarised in \mbox{tables \ref{tab:baseline}-\ref{tab:decisiontree}} showing average one-step prediction accuracy and average score per level set.
The baseline values indicate the varying difficulty of the simple test level and the 10 complex problems.
In case of the simple test level, the measured difference in playing performance using the true and the static FM is much larger than in the hard test levels.
The small gap between the average score in the harder levels highlight the complexity of used test levels and may indicate that the heuristic nature of the RHEA agent may not be perfectly suitable for this game.

The evaluation of our proposed learning methods shows that they are capable of predicting upcoming states with very high accuracy.
However, the analysis of the agents' playing performance using a trained model shows that the remaining errors have a strong influence on the agent behaviour, as an increase in accuracy often means an improved playing performance.
The trained model with the highest accuracy (Decision Tree, cross pattern, span=2) achieved the highest average score in both test level sets.
In contrast, the achieved average score using the Hash Map model is comparable to the playing performance when using a static FM.
Overall, our results indicate that our prediction accuracy measure may not yield a fine-grained enough indication of model performance due to the large number of static tiles per transition.

\subsection{Effects of Imperfect Models }

\begin{figure}[!t]
	\begin{center}
	\includegraphics[width = .15\textwidth]{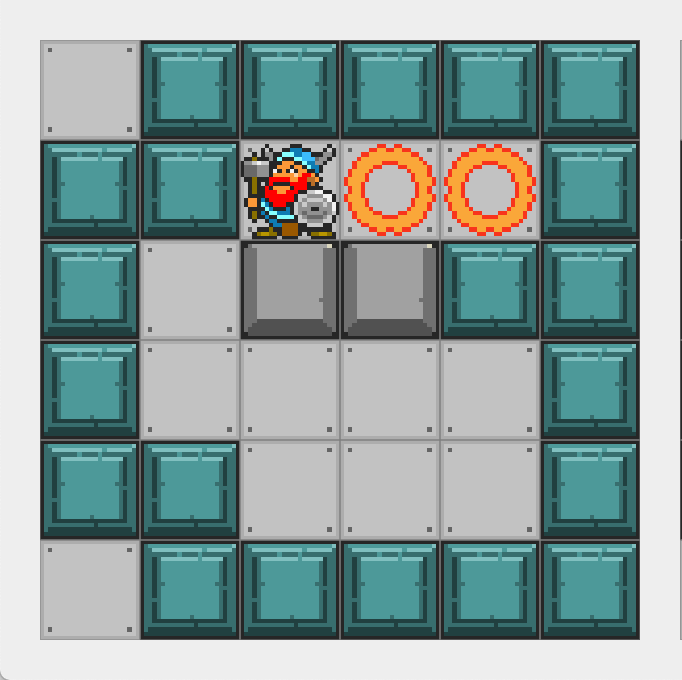}
	\hspace{1em}
	\includegraphics[width = .30\textwidth]{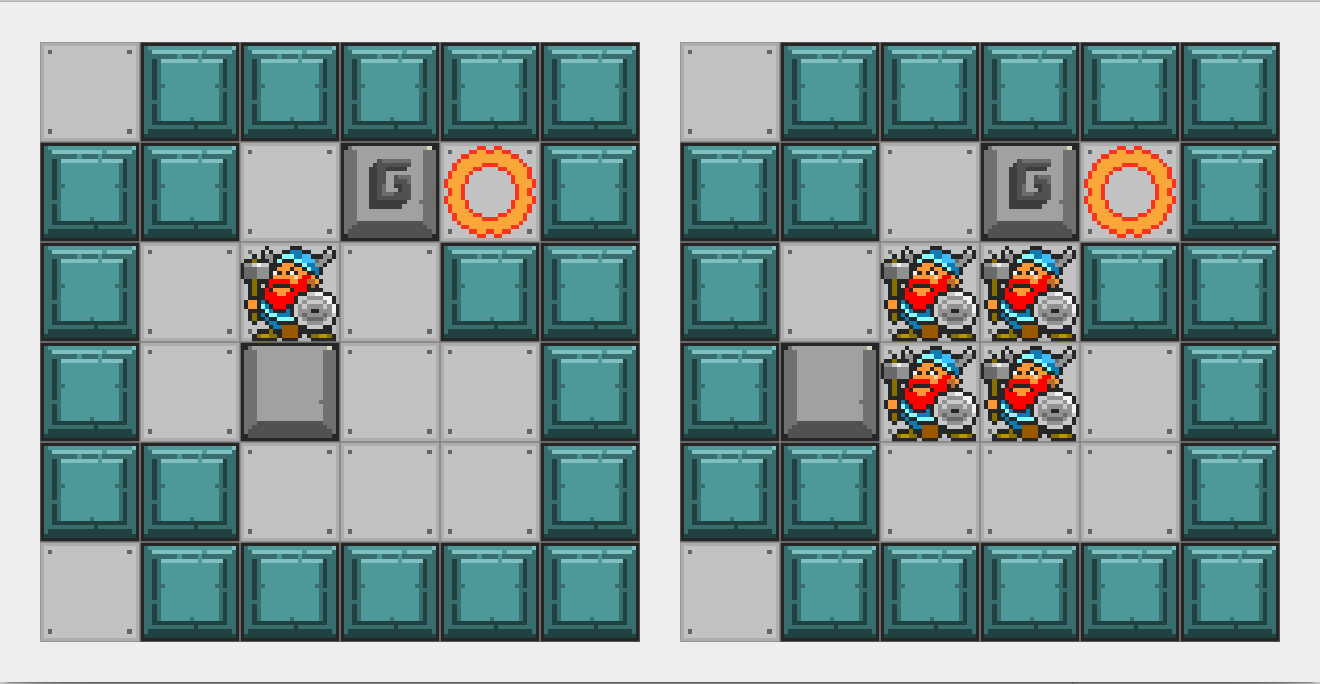}
	\caption{Results of  an  imperfect  model. (left) initial
    game state. (middle) true models' intermediate state en route to the solution, and (right) the estimated state by an imperfect learned model.
    This is after the player has executed the actions: \mbox{DLDRDRRUL-UL},
	where the hyphen denotes the first point at which the learned model diverged from the true model.}
	\label{fig:ModelDivergence}
	\end{center}
\end{figure}

Figure~\ref{fig:ModelDivergence} illustrates how a model can be imperfect but useful.  
The figure shows the true and estimated states 11 steps into a solution to the puzzle shown on the left. 
Note that the estimated model was true to the original one for the first 9 steps, and while it diverged on the 10th step, it still correctly predicted the first "goal" score, with
a box on the left of the two targets.  
After the 11th move, the learned model incorrectly indicates an unsolvable game, with the remaining box stuck in the corner and a proliferation of avatars.  
Nonetheless, an SFP algorithm following this sequence would still have scored a point.

\section{Conclusions and Future Work}
\label{sec:conclusion}

In this paper we tested the recently proposed FM learning approach on Sokoban, which is an unforgiving game (a single wrong action / prediction can lead to unsolvable game states).
While all trained models predicted following game states with very high accuracy, we could show that this is not sufficient to play the game with high performance using the learned model as replacement for the true FM.
Decisions based on wrongly predicted game states may yield non-optimal actions leading the agent into a trap state.
Thus, single errors can have a drastic effect on the resulting playing performance.

In our tests, the Hash Map model was able to predict game states of both the training and the test set with high accuracy.
However, comparing the playing performance of trained Hash Map models to a static FM indicates that the trained models did not yield significantly better results.
In comparison, the best performing Decision Tree model, which is still worse than using the true FM, yielded a significant improvement over using the static FM.
Nevertheless, the game playing performance of the Decision Tree model was shown to be strongly dependent on the used local neighbourhood.
Increasing the number of observed tiles also increases the number of observable patterns and makes it harder to identify important components of the input pattern. This reduces the model's ability to generalise.

Overall, the local learning approach seems to be a promising for cross level learning. As in the experiments in this paper, the agent can be trained on a set of simple levels and the learned model could later be applied to unknown levels.


The work presented in this paper only represents a fraction of the possible directions that this research can take. Using the same experimental settings, it is possible to investigate not only future states, but also rewards learned by the FM. Similarly, varying the shape or span of local patterns would allow to fully test the learning capabilities of the system. Alternatively, dependency analysis could be used to create arbitrarily shaped \mbox{patterns \cite{DockhornCI2018}}. Exploiting rotation invariant patterns may thus prove useful to reduce the number of necessary training samples.

Furthermore, it would be interesting to investigate different similarity measures between game states and the resulting forward model prediction accuracy. This could help to explain the agent behaviour and resulting score better, as well as to identify patterns in Sokoban puzzles.

Future work can also answer interesting research questions that concern the generality of the system. Each level and agent provides a subset of the possible patterns that can be observed in the game. These patterns are the input of the FM learning methods, which can therefore be inaccurate in function of the patterns discovered at the time of training. It is worthwhile asking how many patterns (or levels) are required to train the system so it can be used efficiently for new test levels.
Similarly, repair operators could be used to either repair unrealistic game states (i.e. removing multiple agents as shown figure \ref{fig:ModelDivergence}) or to repair or retrain the model based on new observations.

Here we used NTBEA to tune the agent when using a perfect model, but it may be possible to
extract higher performance from an imperfect model by optimising the RHEA parameters when using that model.

Finally, it is also worth considering other games this work can be extended to, such as Tetris or other GVGAI games.

%
%
%
%

\bibliographystyle{abbrvnatShort}
{\small
\bibliography{lfm}}

\end{document}